\documentclass[runningheads]{llncs}
\usepackage{graphicx}
\usepackage{amsmath,amssymb} 
\usepackage{color}
\usepackage{subfigure}
\usepackage{booktabs}

\begin{document}
	\pagestyle{headings}
	\mainmatter
	\def\ECCVSubNumber{???}  
	
	\title{SDOD: \ Real-time Segmenting and Detecting 3D Object by Depth} 

	\titlerunning{SDOD: \ Real-time Segmenting and Detecting 3D Object by Depth}
	%
	\author{Shengjie Li\inst{1} \and
		Caiyi Xu\inst{1} \and
		Jianping Xing\inst{1} \and
		Yafei Ning\inst{1} \and
		YongHong Chen\inst{2}}
	\authorrunning{Li, S. et al.}
	%
	\institute{School of Microelectronics, Shandong University, Jinan, China \and
		City Public Passenger Transport Management Service Center of Jinan, China 
		\email{lishengjie199012@126.com, cyxu@mail.sdu.edu.cn, \{xingjp, ningyafei\}@sdu.edu.cn, chenyhjn@foxmail.com}}
	\maketitle
	
	\begin{abstract}
		Most existing instance segmentation methods only focus on improving performance and are not suitable for real-time scenes such as autonomous driving. This paper proposes a real-time framework that segmenting and detecting 3D objects by depth. The framework is composed of two parallel branches: one for instance segmentation and another for object detection. We discretize the objects' depth into depth categories and transform the instance segmentation task into a pixel-level classification task. The Mask branch predicts pixel-level depth categories, and the 3D branch indicates instance-level depth categories. We produce an instance mask by assigning pixels which have the same depth categories to each instance. In addition, to solve the imbalance between mask labels and 3D labels in the KITTI dataset, we introduce a coarse mask generated by the auto-annotation model to increase samples. Experiments on the challenging KITTI dataset show that our approach outperforms LklNet about 1.8 times on the speed of segmentation and 3D detection.
		
		\keywords{Real-time Instance Segmentation, 3D Object Detection, Depth Estimation}
	\end{abstract}

	\section{Introduction}
	
\begin{figure}[htpb]
		\centering
		\includegraphics[width=1.0\textwidth]{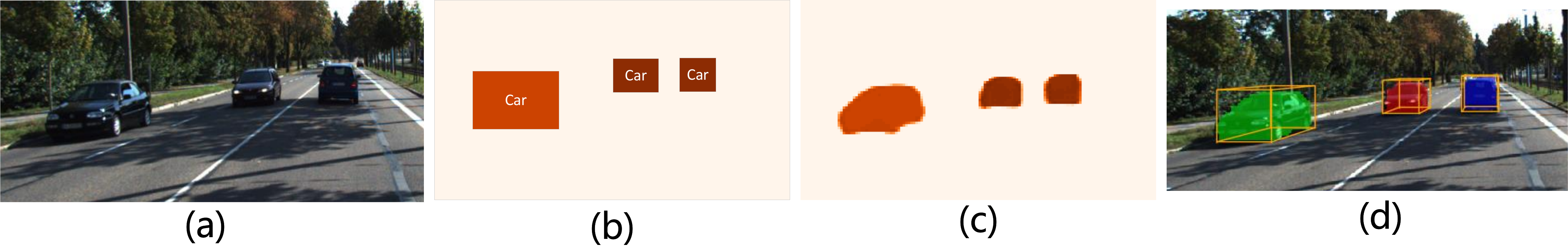}
		\caption{\textbf{Input and output} Fig.(a) input of SDOD; Fig.(b) instance-level depth categories map; Fig.(c) pixel-level depth categories map; Fig.(d) fused instance segmentation and detection output. In Fig (b) and (c) darker the color of a pixel/instance is, the greater the depth value of a pixel is, and the farther the pixel/instance is from us.}
		\label{Four}
	\end{figure}

	With the development of automated driving technology, real-time instance segmentation and 3D object detection of RGB images play an increasingly important role. Instance segmentation combines object detection and semantic segmentation, help autonomous vehicles perceive complex surroundings. State-of-the-art approaches to instance segmentation like Mask R-CNN  \cite{he2017mask} and Fully Convolutional Instance-aware Semantic Segmentation (FCIS) \cite{li2017fully} are only focused on 2D objects and are still show poor performance in 3D scenes such as inaccurate semantic masks or error labels. Therefore, it is not enough to be used in autonomous driving. For more accurate 3D locations and shorter inference time, some latest 3D object detection frameworks speed up by splitting 3D tasks into multiple 2D related subtasks, like MonoGRNet \cite{qin2019monogrnet} proposes a network composed of four task-specific subnetworks, responsible for 2D object detection, instance depth estimation, 3D localization and local corner regression. In this paper, we focus on the real-time task fusion of instance segmentation and 3D object detection. Different from 3D instance segmentation based on the point cloud, We propose a framework whose 3d branches and mask branches are parallel and proposal-free; when we input an image, it can output 3D location, 3D bounding box and instance mask by depth in real-time, as shown in Fig. \ref{Four}.
	
	Object detection is the foundation of computer vision. 2D object detection methods based on convolutional neural networks (CNN) \cite{krizhevsky2012imagenet} such as Faster R-CNN \cite{ren2015faster}, You Look Only Once (YOLO)  \cite{redmon2016you}, Single Shot MultiBox Detector (SSD) \cite{liu2016ssd} have achieved highly accuracy. The methods above are anchor-based, Fully Convolutional One-Stage Object Detection (FCOS) \cite{tian2019fcos} uses no anchor and performance better than anchor-based methods. Multinet  \cite{teichmann2018multinet} proposes a non-proposed approach similar to YOLO and uses RoiAlign \cite{he2017mask} in the rescaling layer to narrow the gap. Existing methods based on RGB images include multi-view method Multi-View 3D (MV3D) object detection network \cite{chen2017multi}, single-view method MonoGRNet, MonoFENet \cite{bao2019monofenet} and RGB-Depth method. MV3D takes the bird's eye view and front view of the point cloud and an image as input; the RGB-Depth method takes RGB image and point cloud depth as input; MonoGRNet takes a single RGB image as input and proposes a network composed of four task-specific subnetworks, responsible for 2D object detection, instance depth estimation, 3D localization and local corner regression. Depth estimation is mainly divided into monocular depth estimation and binocular depth estimation\cite{kendall2017end}. DORN \cite{fu2018deep} proposes a spacing-increasing discretization method to discretize continuous depth values and transform depth estimation tasks into classification tasks. This method estimates each pixel's depth in the image; it may not be suitable for 3D object detection.
	
	Existing methods range from one-stage instance segmentation approach YOLACT \cite{bolya2019yolact++}, Segmenting Objects by Locations (SOLO) \cite{wang2019solo} to two-stage instance segmentation approach Mask R-CNN, Mask Scoring R-CNN \cite{huang2019mask}. Mask R-CNN is a representative two-stage instance segmentation approach that first generates ROI(region-of-interests) and then classifies and segments it in the second stage. Mask Scoring R-CNN as an improvement on Mask R-CNN adds a new branch to score the mask to predict a more accurate score. Like two-stage object detection, two-stage instances are based on the proposal. They perform well but slowly. SOLO distinguish different instances by 2D location. It divided an input image of $H \times W$ into $S_{x}\times S_{y}$ grids and do semantic segmentation in each grid; this is similar to the main idea of YOLO. However, it only uses 2D locations to distinguish different instances, performance not good for overlapped instances. Therefore, we propose to solve the task by tackling three issues: 1) how to transform instance segmentation tasks into semantic segmentation tasks, 2) how to combine the 3D network with the instance network efficiently, 3) how to train the 3D network and the instance network together.
	
	\begin{figure}[htpb]
		\centering
		\includegraphics[width=0.9\textwidth]{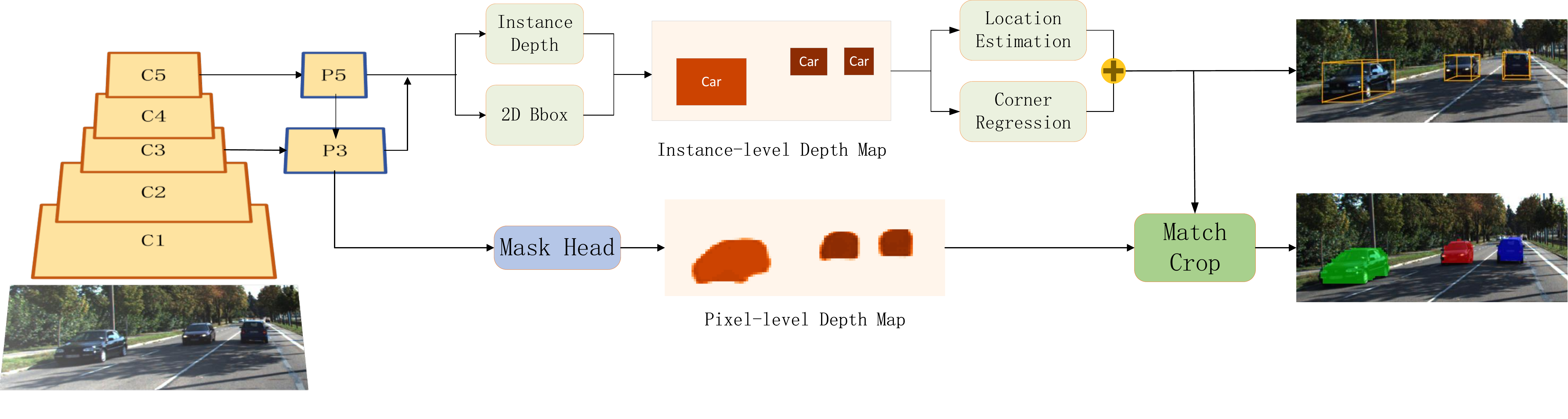}
		\caption{\textbf{SDOD Framework.} SDOD consists of a backbone network and two parallel branches: a 3D branch and a mask branch. We match and crop the instance-level depth category map generated by the 3D branch and the pixel-level depth category map generated by the mask branch. Finally, we will get an instance mask.
		}
		\label{framework}
	\end{figure}  
	
	For these problems, we use depth to connect the 3D network with the instance network, and at the same time, use depth to transform instance segmentation into semantic segmentation. As shown in Fig. \ref{framework}, the network has been divided into two parallel branches: the 3D branch and the mask branch. The objects' depth discretized into depth categories, the 3D branch predicts instance-level depth categories, and the mask branch indicates pixel-level depth categories. We introduce the auto-annotation model trained on Cityscapes provided by Polygon-RNN++ \cite{acuna2018efficient} to generate coarse masks on the KITTI dataset \cite{geiger2012we}. Then add real depth to these coarse masks. Finally, we use these masks to train the mask branch. Experiments on the  KITTI dataset demonstrate that our network is practical and real-time.
	
	In summary, in this paper we propose a framework for real-time instance segmentation and 3D object detection. For the time to complete all tasks, it can outperform the state-of-the-art about 1.8 times. Our contributions are three-fold:
	\begin{itemize}
		\item Transform instance segmentation tasks into semantic segmentation tasks by discretion depth. 	
		\item Propose a network that combines 3D detection and instance segmentation and set them as parallel branches to speed up.	
		\item Combine coarse masks with real depth to train the mask branch to solve imbalanced labels.
	\end{itemize}
	
	\section{Materials and Methods}
	\subsection{Dataset}
	
	KITTI dataset has 7841 training images and 7581 test images with calibrated camera parameters for 3D object detection challenges. However, due to the difficulty of instance segmentation labeling, there are only 200 labeled training images and 200 unlabeled testing images for instance segmentation challenge. In addition, the 3D object detection task evaluates on 3 types of targets(car, pedestrian, cyclist), and instance segmentation task evaluates on 8 types of targets(car, pedestrian, cyclist, truck, bus, train, bicycle, motorcycle).We evaluate cars, pedestrians, and cyclists on both 3D object detection and instance segmentation tasks.
	
	The number of 3D objects in the KITTI dataset is slightly less than the number of instance masks. 3D detection dataset ignores targets beyond the Lidar detection range, and some of them are not forgotten in the instance segmentation dataset. We take the 3D dataset as the benchmark in our work, although this will bring some performance loss.
	
	\subsection{Instance Segmentation}
	
	It is hard to directly regress the continue center depth $g_{d}$, we discretize continuous depth into depth classes, and a particular class $c_i$ can be assigned to each depth $d$. There are two discrete methods: linear method and non-linear method. Linear method means that the depth $d\in[d_{min},d_{max}]$ is linearly divided into classes $c_i\in\left\{c_1,c_2,...,c_{K}\right\}$. Note that the background is set to $c_0$ and the value is $-1$. Non-linear method chooses a more complex mapping function for discretization e.g. SDNet  \cite{ochs2019sdnet} chooses a logarithmic function and DORN chooses an exponential function. 
	
\begin{figure}[htpb]
		\centering
		\includegraphics[width=1\textwidth]{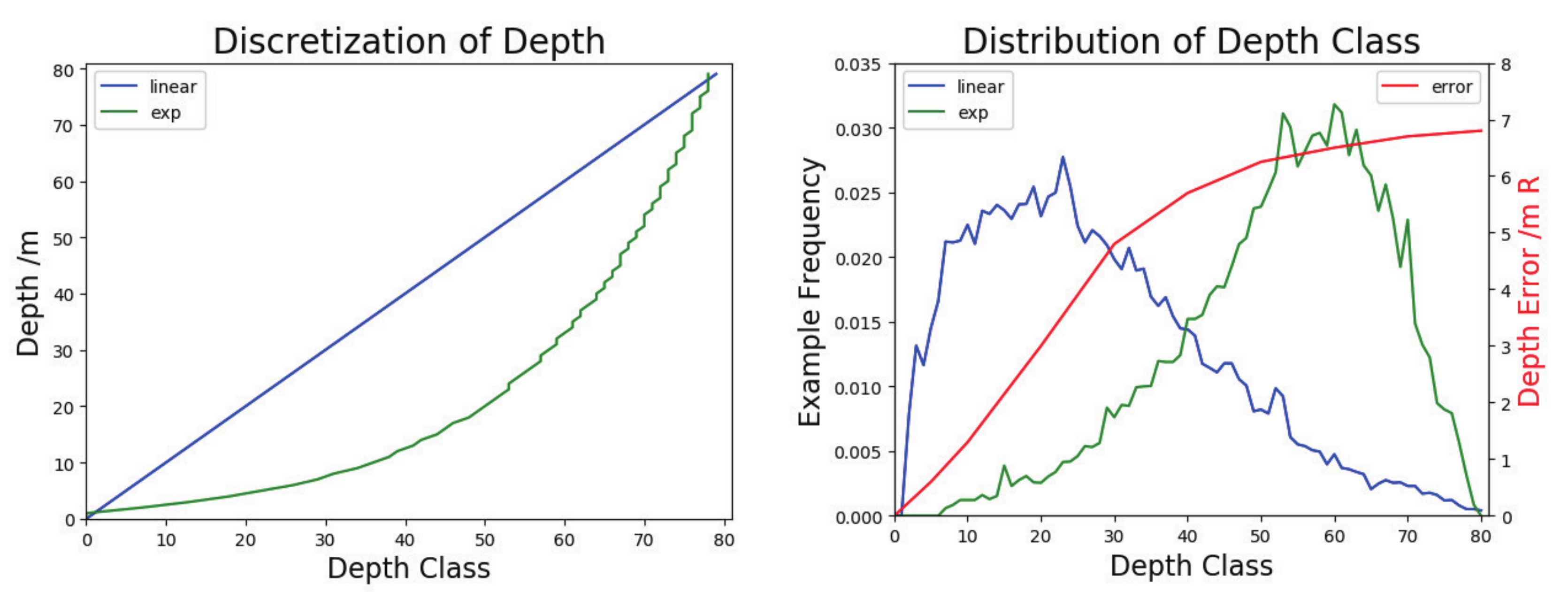}
		\caption{The left plot shows the linear and exponential discretization of the depths with $K=80$ in a depth interval [2, 80]. The right plot shows the example frequency of linear and exponential discrete depth classes in the KITTI 3D object detection dataset. The depth error of example in different depth classes is also shown in the right plot. The depth error curve reflects the difficulty of the sample, and the data comes from 3DOP \cite{chen20153d}.}
		\label{SID}
	\end{figure}

	Compared with the discrete linear method, the non-linear discrete method increases the proportion of difficult examples, making the model easier to train and converge. In this work, we spilled the depth into depth classes $c_i$ with an exponential function \ref{dorn} where K is the number of depth classes.
	\begin{equation}\label{dorn}
	c_i = d_{min}\cdot\left(\frac{d_{max}}{d_{min}}\right)^{\frac{i-1}{K-1}} \ , i\in\left\{1,2...,K\right\}
	\end{equation}
	The left plot of Fig. \ref{SID} shows the linear and exponential discretization of the depths, the right plot shows the example frequency of linear and exponential discrete depth classes in the KITTI 3D object detection dataset. Simultaneously, we use depth error to measure the difficulty of the depth classes, and the red curve shows the depth class of the object is positively related to the difficulty of the object depth estimation. The discrete exponential method increases the proportion of hard examples, making the model easier to train and converge.
	
	\subsection{3D Branch}
	
	We leverage the design of MonoGRNet, which decomposes the 3D object detection into four subnetworks: 2D detection, instance-level depth estimation, 3D location estimation and corner regression.
	
	We use the design of 2D detection in Multinet, which proposes a non-proposed approach similar to YOLO and Over feat \cite{sermanet2013overfeat}. To archive the good detection performance of proposal based detection systems, it uses RoiAlign in the rescaling layer. An input image of $H \times W$ has divided into $S_{x}\times S_{y}$ grids, and each grid is responsible for detecting objects whose center falls into the grid. Then each grid outputs the 2D bounding box $B_{2d}$ and the class probabilities $P_{cls}$.
	
	Given a grid $g$, this module predicts the center depth $g_{d}$ of the object in $g$ and provides it for 3D location estimation and mask branch. As shown in Fig. \ref{framework}, the module takes P5 and P3 as the input feature map. Compared with P3, P5 has a larger receptive field and lower resolution. It is less sensitive to location, so we use P5 to generate a coarse depth estimation, and then fused with P3 to get accurate depth estimation. We apply several parallel atrous convolutions with different rates to get multi-scale information, then fuse it with a 2D bounding box to generate an instance-level depth map.Compared with the mask branch's pixel-level depth estimation, the module output resolution is lower, which is an instance-level. For details of implementation, please refer to section 2.4.
	
	The 3D location module uses the 2D coordinates $ \left(u,v\right)$ and the center depth $d$ of the object to calculate the 3D location $\left(x,y,z\right)$ by the following formula:
	\begin{equation}
	\left\{
	\begin{array}{lr}
	u = x\cdot f_{x} + cx &  \\
	v = y\cdot f_{y} + cy\\
	d = z &  
	\end{array}
	\right.
	\end{equation}
	$f_x, f_y, cx, cy$ are camera parameters which can be obtained from the camera's internal parameter matrix C.
	
	As illustrated in Fig.\ref{sigma}, we first establish a coordinate system whose origin is the object center, and the $x$ ,$y$ ,$z$ axis is parallel to the camera coordinate axis, and then regress the 8 corners of the object. Finally, we use the method of Deep3DBox \cite{mousavian20173d} to calculate the object's length, width, height, and observation angle from 8 corner points. The length, width and observation angle will be used to calculate the depth threshold in Section 2.5.
	
	\subsection{Mask Branch}
	The mask branch predicts pixel-level depth categories over the entire image, and classifies pixels based on the depth class to which they belong, which is similar to semantic segmentation. As shown in Fig. \ref{ASPP}, the mask branch consists of
	atrous spatial pyramid pooling(ASPP) layers, fully convolutional (FCN) layers, fully connected (FC) layers, and upsample layers. ASPP layers help to get multi-scale information, FCN layers help to get semantic information, and FC layers help to transform semantic information into depth information. We have tried using convolutional layers instead of FC layers and encoding the depth category, but the performance is not good.
	
	\unskip
	\subsubsection{ASPP}
	
	The ASPP module's input is the P5 feature map, and its resolution is only 1/32 of the original image. To expand the receptive field of the input and obtain more semantic information, we use the ASPP module, inspired by dilated convolutions \cite{yu2016multi} and DeepLab v3++ \cite{chen2018encoder}. 
	
\begin{figure}[htpb]
		\centering
		\includegraphics[width=0.9\textwidth]{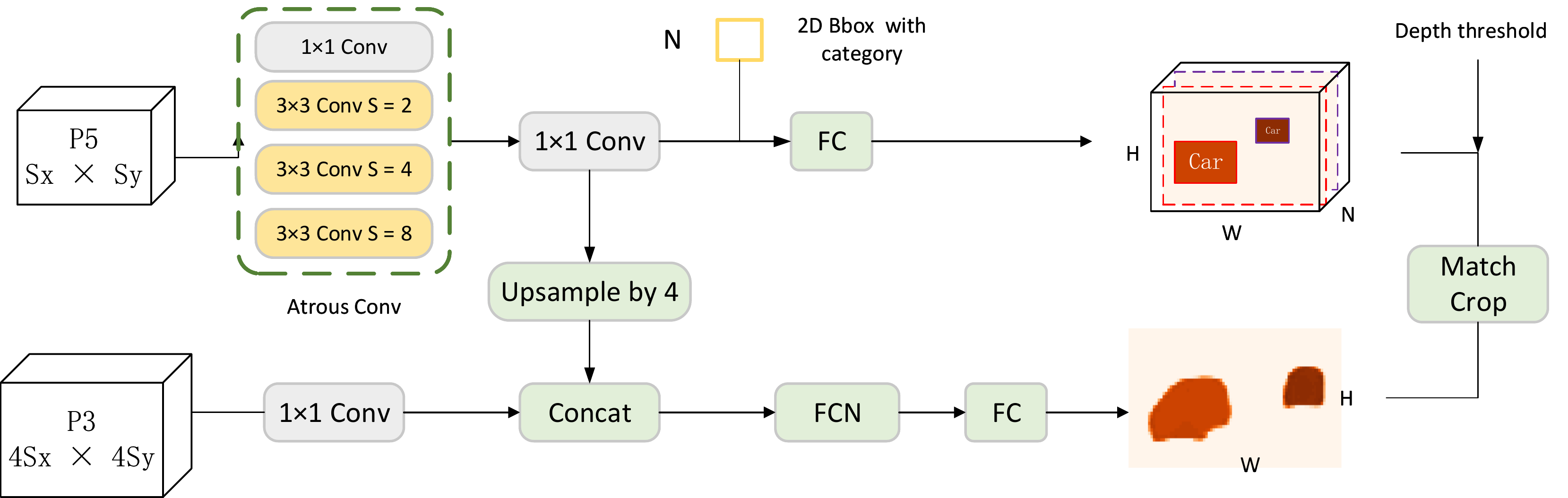}
		\caption{Mask branch architecture. \ The mask branch consists of ASPP layers, FCN layers, FC layers, and upsample layers. Note that the height $H$ and width $W$ in the picture is 1/4 of the original input picture. }
		\label{ASPP}
	\end{figure}

	As shown in Fig. \ref{ASPP}, The ASPP module connects 1 convolutional layer and 3 atrous convolutional layers with rates of 2,4,8. The module's input size is 39$\times$12$\times$512; after upsampled and concatenated, it becomes to 156$\times$48$\times$256, then we throw the feature map into FCN layers. 
	
	\unskip
	\subsubsection{FCN and FC}
	
\begin{figure}[htpb]
		\centering
		\includegraphics[width=0.9\textwidth]{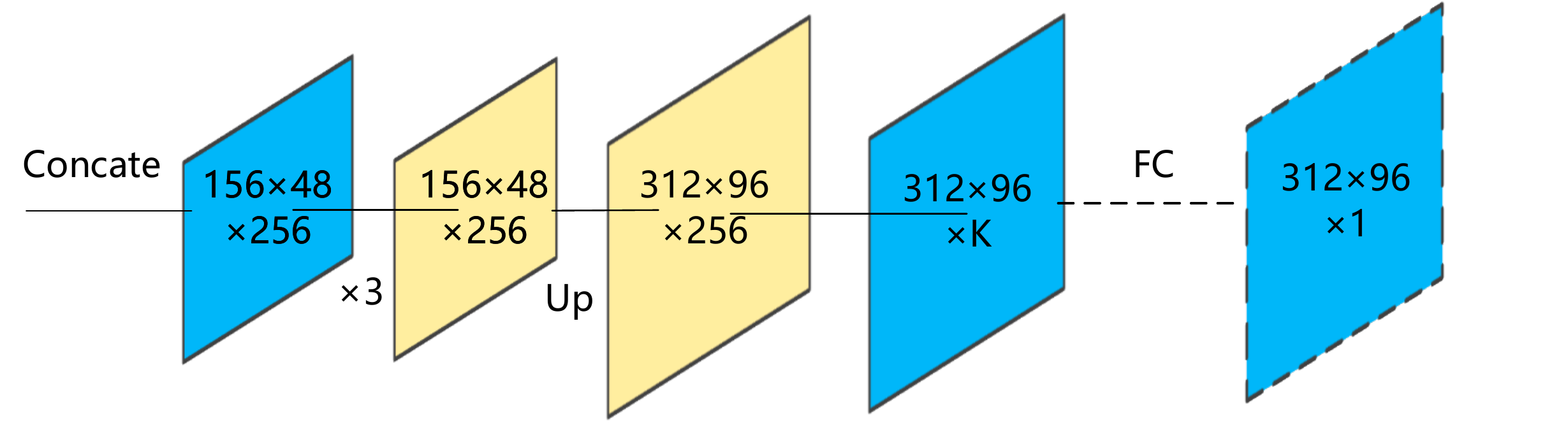}
		\caption{\textbf{FCN with FC} \ The brown feature map is obtained by 3$\times$3 conv, and the blue feature map is obtained by 1$\times$1 conv. $\times$3 means that 3 conv layers are used. Up means that the upsampling layer is used, and K is the total number of depth categories. We apply fully connected layers to get the exact value of the depth category that varies from 0 to K(background is 0). Note that the mask branch does not predict the pixel's target category (car, pedestrian, cyclist). It is indicated in the 3D branch. }
		\label{FCN}
	\end{figure}

	To get a pixel-level depth category map, we use an FCN module similar to the mask branch in Mask R-CNN, proposed by FCN \cite{long2015fully}. As shown in Fig \ref{FCN}, compared to Mask R-CNN, we have added a 1$\times$1 convolution layer, which is responsible for the depth classification of each pixel. K is the total number of depth categories in equation \ref{dorn}; we set it to 64. The Mask branch does not predict the pixel's target category (car, pedestrian, cyclist); it is expected in the 3D branch. The FCN module finally outputs 1 pixel-level depth category map, as shown in Fig. \ref{ASPP}. The darker the pixel's color, the greater the pixel's depth value and the farther the pixel is from us. The size of the output image is 312$\times$96, and the size of the original image is 1248$\times$384.
	
	\subsubsection{Coarse Mask Generation}
	
	To solve the imbalanced between mask labels and 3D labels in the KITTI dataset, we introduce a coarse mask generated by the auto-annotation model to increase instance segmentation samples. Polygon-RNN++ is a state-of-the-art auto-annotation model which inputs 2D bounding boxes and outputs instance masks. It is trained on the instance-level semantic labeling task of the Cityscapes dataset. We use 200 labeled training images to evaluate the accuracy of the coarse mask. Results showed in Table \ref{coarse}. 
	
	We did not directly train the mask branch with coarse labels , but superimposed the real depth value with it for training:
	\begin{equation}\label{trans}
	p_k=i_k\times m_{k},  \  i_k\in\left[1,K\right], \ m_{k}\in\left\{0,1\right\}
	\end{equation}
	$i_k$ is the real depth category of instance $k$, which can be calculated from equation \ref{dorn}. $m_{k}$ is the coarse mask of instance $i$, with a value of 0 or 1, $p_{k}$ is the final label for mask branch. When training the mask branch, first, we train the mask branch and the 3D branch together with coarse masks for 120K iterations, and then train the mask branch only with fine masks for 40K iterations.
	\begin{table}[htbp]
		\centering
		\caption{Accuracy of coarse mask generated by Polygon-RNN++. Note that Polygon-RNN++ was trained on the Cityscapes dataset rather than the KITTI dataset.}
		\begin{tabular}{ccccc}
			\toprule[1pt]  
			& \ car & \ pedestrian & \ cyclist & \ average \\ 
			\midrule[1pt]  
			AP & \ 40.1 & \ 36.3 & \ 35.1 & \ 37.2 \\
			$AP_{50}$ & \ 56.7 & \ 50.6 & \ 50.3 & \ 52.5 \\
			\bottomrule[1pt]  
		\end{tabular}
		\label{coarse}
	\end{table}

	\subsection{Match And Crop}
	Instance segmentation requires each pixel to be assigned to a different instance. We need to assign each pixel in the pixel-level depth map $X = \left\{x_0,x_1,x_2...x_{N-1}\right\} $ to a set of instance $S = \left\{S_0,S_1,S_2...S_{M-1}\right\}$ in the 3D branch,and we treat this as a pixel matching task. 
	
	How to match pixels with instances? There are two conditions: first, the pixel must have the same depth category as the instance; second, the pixel must have the same position as the instance. The following formula can describe the first condition:
	\begin{equation}\label{cluster}
	x_i\in S_k\Leftrightarrow\mid x_i-S_k\mid < \delta_k, \ i\in\left[0,N-1\right], \ k\in\left[0,M-1\right]
	\end{equation}
	$x_i$ is the depth class of pixel $i$ and $x_i\in\left[0,K\right]$, $S_k$ is the depth class of the instance $k$, and $\delta_k$ is the depth threshold of the instance $k$.
	
	As shown in Fig.\ref{sigma}., each instance has only one depth class in the instance-level depth map, but each instance may has multiple depth classes in the pixel-level depth map. So we set a depth threshold $\delta_k$ for each instance, which is calculated by the following formula:
	\begin{equation}\label{sigma}
	\begin{split}
	\delta_k = \left(K-1\right)\cdot\log_{d_{max} / d_{min}}{\frac{c_k}{c_k-\triangle d_k}}
	\end{split}
	\end{equation}
	\begin{equation}\label{delta}
	\begin{split}
	\triangle d_k=\frac{1}{2}w_k\mid\cos\theta_k\mid+\frac{1}{2}l_k\mid\sin\theta_k\mid,\quad\theta_k\in\left[-\pi,\pi\right]
	\end{split}
	\end{equation}
	$c_k$ is the depth class of instance $k$, $\delta_k$ is the depth margin and is shown in Fig.\ref{sigma}. $w_k$, $l_k$ and $\theta_k$ are the width length  and observation angle of the instance, which can be obtained from the cornesr regression module. The derivation of equation \ref{delta} can be seen in the supplementary material.
	
	The second condition can be transformed into a crop operation. Crop operation means using the 2D bounding box to crop the pixel-level depth map, improving the mask's accuracy. During training, we use truth bounding boxes to crop the depth map.
	
	\begin{figure}[htpb]
		\centering
		\includegraphics[width=0.30\textwidth]{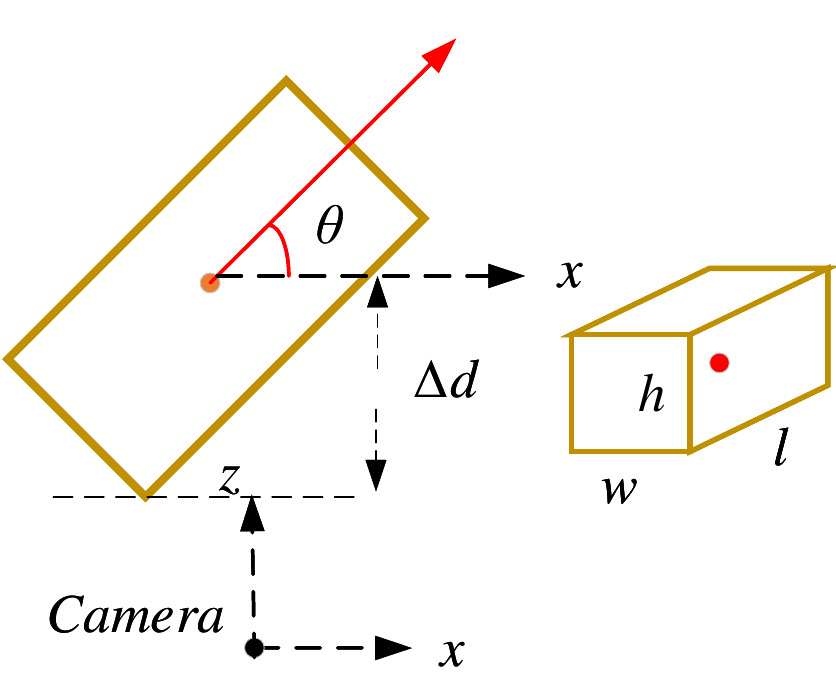}
		\caption{Coordinate system and geometric constraints. The left one is a bird's eye view and shows the position of the target and the camera. $\triangle d$ is the depth threshold and can be calculated by equation \ref{delta} , $\theta$ is the deflection angle of the object. }
		\label{sigma}
	\end{figure}
	
	\subsection{Loss Function}
	Here we have determined independent loss functions for each module and joint loss functions for the entire network. 2D detection includes classification loss $L_{cls}$ and box regression loss $L_{box}$, they are defined in the same as in Multinet. Due to the imbalance of samples between classes in the KITTI dataset, we used focal loss \cite{lin2017focal} to modify the classification loss. We use L1 loss as instance-level depth loss, corner loss and location loss, and they are the same as in MonoGRNet. When fully connected layer is used, pixel-level depth loss is the same as the instance-level depth loss.
	\begin{equation}\label{2D}
	L_{2d} = w_1L_{cls} + w_2L_{box}
	\end{equation}
	The total loss of 2D inspection. Where $w_1$ and $w_2$ are weight coefficients, when we trained 2D detection only, we set  $w_1=w_2=1$.
	
	\begin{equation}\label{inst-depth}
	L_{d} = \sum_{i=1}^n\mid d_i-\hat{d_i}\mid
	\end{equation}
	Instance-level depth loss. Where n is the numbel of cell, $di$ is the ground truth of cell $i$, $\hat{d_i}$ is the prediction of cell $i$.
	
	\begin{equation}\label{pixel-depth}
	L_{mask} = \sum_{i=1}^n\mid M_i-\hat{p_i}\mid
	\end{equation}
	Pixel-level depth loss. Where n is the numbel of pixel, $M_i$ is the ground truth of pixel-level depth categories, which is deined by equation  \ref{trans} , $p_i$ is the prediction category of pixel $i$.
	We also tried L2 loss, CE loss and Focal loss, and finally we found that L1 loss performed better. We think that the smaller the object is, the farther it is, the greater its depth value is and the greater the loss is. Moreover, this is why the long-distance object can be detected well.
	
	\subsection{Implement Details}
	
	The architecture of SDOD is shown in Figure \ref{framework}. VGG-16\cite{simonyan2014very} is employed as the backbone but without the FC layers. FPN \cite{lin2017feature} is used to solve the problem of multi-scale detection. The 2D detector should be trained first. We set $w_1=w_2=1$ in the loss functions and initialize VGG-16 with the pre-trained weights on ImageNet. We trained a 2D detector for 150K iterations with the Adam optimizer\cite{kingma2014adam}, and L2 regularization is used with a decay rate 1e-5. Then the 3D branch and the mask branch are trained for 120K iterations with the Adam optimizer. At this stage, coarse mask generated by Polygon-RNN++ is used. Finally, we continue to train the network with fine masks for 40K iterations with an SGD optimizer. We set the batch size to 4, learning rate to 1e-5 and dropout rate to 0.5 throughout the training.

	\section{Results}
	
	The proposed network is developed using Python on a single GPU of NVidia GTX 2080TI. For evaluating 3D detection performance, we follow the KITTI benchmark's official settings to evaluate the 3D Average Precision($AP_{3d}$). For evaluating instance segmentation performance, we follow the official settings of the KITTI benchmark to evaluate the Average Precision on the region level ($AP$) and Average Precision with 50$\%$($AP_{50}$). In this work we only evaluate three types of objects: car, pedestrian, and cyclist. The results show that the car has the highest accuracy and the rider has the lowest accuracy, as shown in Table \ref{detail}. We compared our method with Mask R-CNN and Lklnet \cite{Detectron2018} by evaluating the AP and $AP_{50}$ of instance segmentation tasks, and the results are shown in Table \ref{inst_result}. Though Mask R-CNN has higher accuracy, Our approach is almost 18 times faster than that. Even if compared with the fastest two-stage instance segmentation method, we gain about 1.8 times relative improvement on speed, which is more suitable for autonomous driving. We also evaluated our method on 3D object detection tasks and compared with MonoFENet and  MonoPSR \cite{ku2019monocular}. Thanks to splitting the 3D detection into four sub-networks, we gain around 3.3 to 4.7 times relative improvement on speed, and the results are shown in Table \ref{3D}. 
	
	\begin{table}[htbp]
		\centering
		\caption{Specific accuracy for each category of our method. Note that our method only evaluate AP and $AP_{50}$ of car, pedestrian, and cyclist.}
		\begin{tabular}{ccccc}
			\toprule[1pt]  
			& \ car & \ pedestrian & \ cyclist & \ average \\ 
			\midrule[1pt]  
			AP & \ 23.36 & \ 19.73 & \ 18.15 & \ 20.38 \\
			$AP_{50}$ & \ 48.59 & \ 33.21 & \ 30.26 & \ 37.35 \\
			\bottomrule[1pt]  
		\end{tabular}
		\label{detail}
	\end{table}
	
	\begin{table}[htbp]
		\centering
		\caption{Instance segmentation mask AP on KITTI. Mask R-CNN is trained on the KITTI dataset and inference with the environment of 1 core 2.5 Ghz (C/C++). All parameters are tuned in COCO dataset in Mask R-CNN*. Ours time is the total time of 3D detection and instance segmentation.}
		\begin{tabular}{cccccc}
			\toprule[1pt]  
			Method & \ Backbone & \ Training  & \ $AP_{50}$ & \ AP & \ Time \\ 
			\midrule[1pt]  
			Mask R-CNN & \ ResNet101+FPN &   KITTI & \ 39.14 & \ 20.26 & \ 1s \\
			Mask R-CNN*  & \ ResNet101+FPN &  COCO & \ 19.86  & \ 8.80   & \ 0.5s \\
			Lklnet & \ ResNet101+FPN &  KITTI  & \ 22.88 & \ 8.05    & \ 0.15s \\
			\textbf{Ours}  & \ VGG16+FPN &  KITTI  & \ 37.35 & \ 20.38 & \ \textbf {0.054s} \\         
			\bottomrule[1pt]  
		\end{tabular}
		\label{inst_result}
	\end{table}
	
	\begin{table}[htbp]
		\centering
		\caption{3D detection performance. All results is evaluated using the $AP_{3D}$ at 0.7 3D IoU threshold for car class. Difficulties are define in KITTI. Ours-3D time only includes 3D inference time and does not include instance segmentation time.  }
		\begin{tabular}{ccccc}
			\toprule[1pt]  
			&Easy &Moderate &Hard & Time \\ 
			\midrule[1pt]  
			MonoFENet & \ 8.35\% & \ 5.14\% & \ 4.10\% & \ 0.15s\\
			MonoPSR & \ 10.76\% & \ 7.25\% & \ 5.85\% & \ 0.20s\\
			\textbf{Ours-3D} & \ 9.63\% & \ 5.77\% & \ 4.25\% & \ \textbf {0.035s}\\
			\bottomrule[1pt]  
		\end{tabular}
		\label{3D}
	\end{table}
	
	\begin{figure}[htpb]
		\centering
		\includegraphics[width=0.9\textwidth]{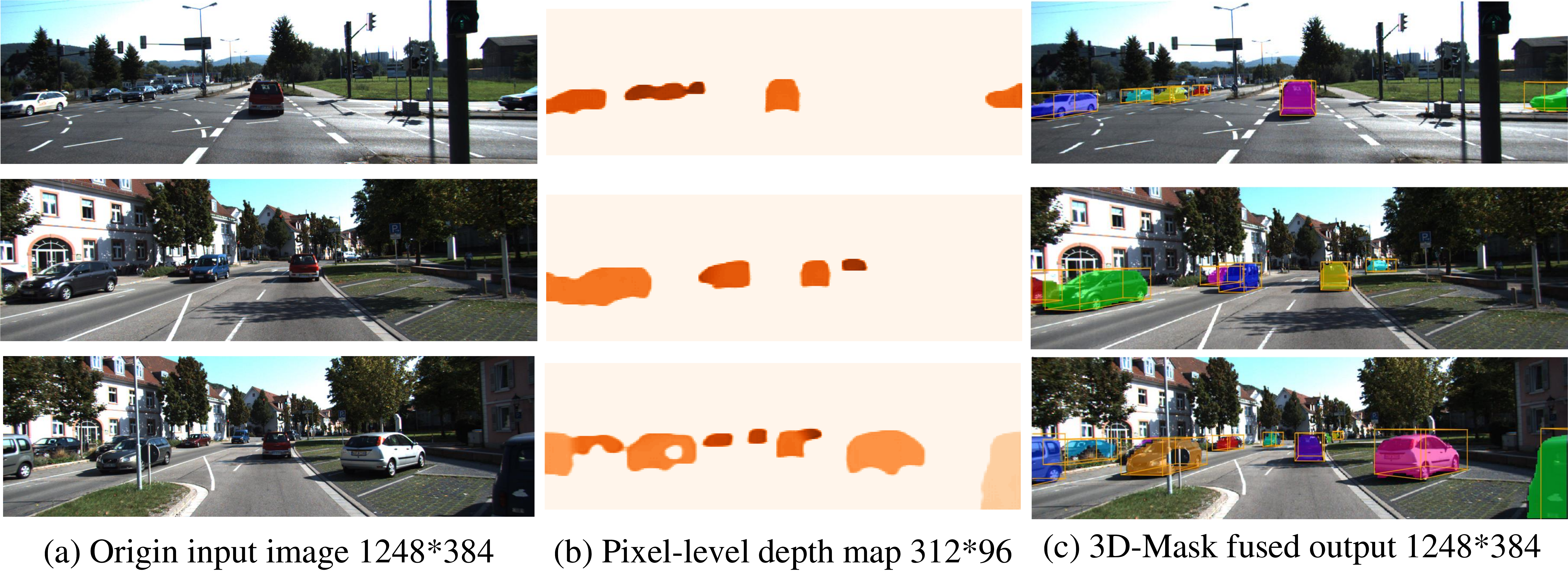}
		\caption{\textbf{Results Samples} Results on KITTI datasets.
			Fig.(a) input of SDOD; Fig.(b) pixel-level  depth categories map; Fig.(c) fused instance segmentation and detection output. We use random colors for instance segmentation. In Fig (b), the darker the pixel's color, the greater the pixel's depth value, and the farther the pixel is from us.}
		\label{All}
	\end{figure}
	
	\section{Discussion}
	\unskip
	
	In the instance-level and pixel-level depth estimation, we use equation \ref{dorn} to discretize the depth into K categories. To illustrate the sensitivity to the number of categories, we set K to different values for comparison experiments, and the results are shown in Table\ref{our_time}. We can see that neither too few nor too many depth categories are rational: too few depth categories cause a large error, while too many depth categories lose discretization.
	
	\begin{table}[htbp]
		\centering
		\caption{Ablation Study Results.  }
		\begin{tabular}{cccccc}
			\toprule[1pt]  
			FC & \ Depth threshold & \ K & \ $AP$ & \ $AP_{50}$  & \ Time \\ 
			\midrule[1pt]  
			\ Y & \ Y & \ \ 32 \  & \ \ 18.90 & \ \ 33.23 & \ \  0.049s \\
			\ Y & \ Y & \ \ 64 \ & \ \ 20.38 & \ \ 37.35 & \ \ 0.054s\\
			\ Y & \ Y & \ \ 96 \ & \ \ 19.84 & \ \ 37.34 & \ \ 0.058s \\   
			\ Y & \ N & \ \ 64 \ & \ \ 19.27 & \ \ 35.84 & \ \ 0.054s\\   
			\ N & \ Y & \ \ 64 \ & \ \ 18.81 & \ \ 32.15 & \ \ 0.052s\\  
			\bottomrule[1pt]  
		\end{tabular}
		\label{our_time}
	\end{table}
	
	We apply fully connected layers to get the exact value of the depth category that varies from 0 to K(background is 0). We try to remove the FC layer and encode the depth category in one-hot form, e.g. if depth category is 3 just output $\left[1,1,1,0,...,0\right]$. The pixel-level feature map size is 312$\times$96$\times$K, and we use pixel-wise binary cross entropy(BCE) loss to replace the L1 loss. The result is shown in Fig.\ref{our_time}. We can see that a fully connected layer is necessary and improves the mask AP from 18.81 to 20.38.
	
	To quantitatively understand SDOD for mask prediction, we perform two error analysis. First, we replace the predicted pixel-level depth map with the ground truth value and coarse value to evaluate the result's mask branch's effect. Specifically, for each picture, we use equation \ref{dorn} to convert the given masks into a pixel-level depth map. As shown in Table \ref{error1}, if we replace the predicted pixel-level depth map with a coarse mask, the AP increase to 35.37; if we replace it with ground truth, the AP increases to 61.18. The results show that there is still room for improvement in the mask branch. Second, we replace the 3D predicted results with 3D ground truth values, including 2D boxes, 3D depth values, and depth thresholds. As reported in Table \ref{error1}, the AP increase from 20.38 to 20.69; this shows that the 3D branch has less effect on the final mask prediction.   
	
	\begin{table}[htbp]
		\centering
		\caption{Error analysis.}
		\begin{tabular}{ccccc}
			\toprule[1pt]  
			&baseline & \ coarse mask & \ gt mask & \ gt 3D \\ 
			\midrule[1pt]  
			AP & \ 20.38 & \ 35.37 & \ 61.18 & \ 20.69 \\
			$AP_{50}$ & \ 37.35 & \ 49.25 & \ 61.42 & \ 37.98 \\
			\bottomrule[1pt]  
		\end{tabular}
		\label{error1}
	\end{table}
	
	\section{Conclusion}
	This paper proposes the SDOD framework to perceive complex surroundings in real-time for autonomous vehicles. Our framework is presented to fuse 3D detection and instance segmentation by depth, split into two parallel branches for real-time: the 3D branch and the mask branch. We combine coarse masks with real depth to train the mask branch to solve imbalanced labels. Our processing speed is about 19 fps, 1.8 times faster than Lklnet and 8 times faster than Mask R-CNN, significantly outperforms existing 3D instance segmentation methods on tasks of segmentation and 3D detection on KITTI dataset.
	\section{Supplementary material: Depth Treshold Derivation}
	In section 3.4 match and crop, we set a depth threshold $\delta_k$ for each instance, which is calculated by the following formula: 
	\begin{equation}\label{sigma_s}
	\begin{split}
	\delta_k = \left(K-1\right)\cdot\log_{d_{max} / d_{min}}{\frac{c_k}{c_k-\triangle d_k}}
	\end{split}
	\end{equation}
	
	\begin{equation}\label{delta_s}
	\begin{split}
	\triangle d_k=\frac{1}{2}w_k\mid\cos\theta_k\mid+\frac{1}{2}l_k\mid\sin\theta_k\mid,\quad\theta_k\in\left[-\pi,\pi\right]
	\end{split}
	\end{equation}
	
	\subsubsection{Depth discretization}
	In section 3.1, we spilt the depth into depth classes $c_i$ with an exponential formula \ref{dorn_s1} where K is the number of depth classes. Transforming formula \ref{dorn_s1} can get formula \ref{dorn_s2}.
	
	\begin{equation}\label{dorn_s1}
	c_i = d_{min}\cdot\left(\frac{d_{max}}{d_{min}}\right)^{\frac{i-1}{K-1}} \ , i\in\left\{1,2...,K\right\}
	\end{equation}

	\begin{equation}\label{dorn_s2}
	i = 1 +\left(K-1\right)\cdot\log_{d_{max}/{d_{min}}}{\frac{c_i}{d_{min}}}
	\end{equation}

	\subsubsection{Depth treshold}
	Then we prove formula \ref{delta_s} with  formula \ref{dorn_s2}. Figure 1 shows the four possible positions of the object corresponding to the value of $\theta$ from $-\pi$ to $\pi$.

	\begin{equation}
	\begin{split}
	\triangle d_k &= AC + AB\\
	&= OA\cdot\mid\cos\theta_k\mid + AD\cdot\mid\sin\theta_k\mid\\
	&= \frac{1}{2}w_k\mid\cos\theta_k\mid+\frac{1}{2}l_k\mid\sin\theta_k\mid
	\end{split}
	\end{equation}

	\begin{equation}
	\begin{split}
	\delta_k &= i_O - i_D\\
	&= 1 +\left(K-1\right)\cdot\log_{d_{max}/{d_{min}}}{\frac{c_{i_O}}{d_{min}}} - (1 +\left(K-1\right)\cdot\log_{d_{max}/{d_{min}}}{\frac{c_{i_D}}{d_{min}}})\\
	&= (K-1)\cdot\log_{d_{max}/{d_{min}}}{\frac{c_{i_O}}{c_{i_D}}}\\
	&= \left(K-1\right)\cdot\log_{d_{max} / d_{min}}{\frac{c_k}{c_k-\triangle d_k}}
	\end{split}
	\end{equation}

	\begin{figure}[htpb]
		\centering
		\includegraphics[width=1.0\textwidth]{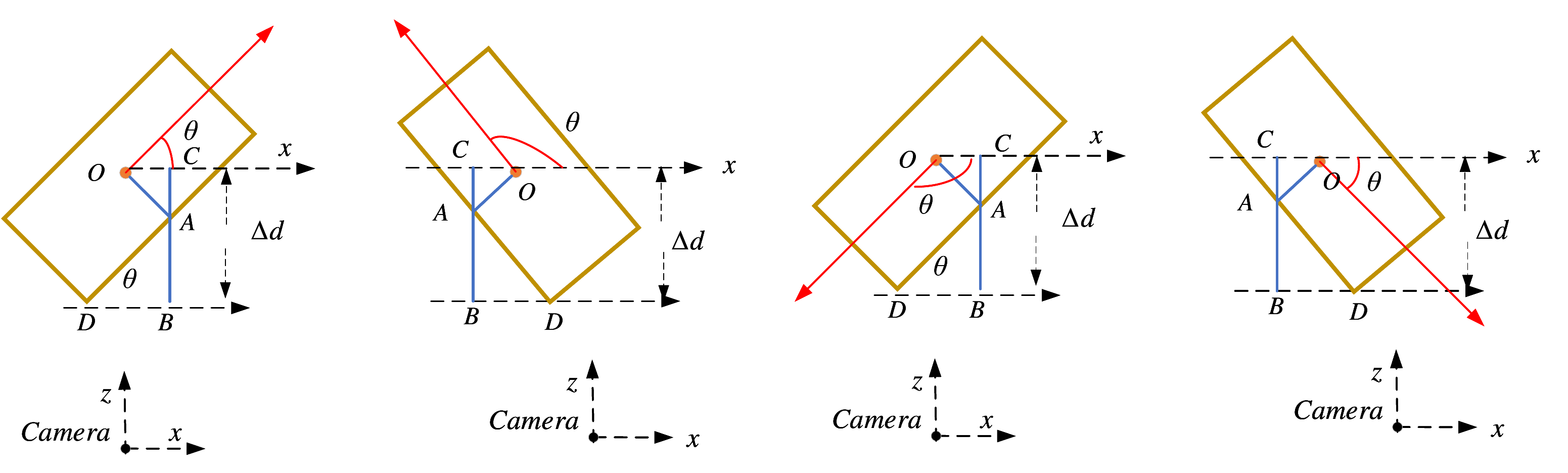}
		\caption{Coordinate system and geometric constraints. It shows the position of the target and the camera. $\triangle d$ is the depth threshold and can be calculated by formula \ref{delta_s} , $\theta$ is the deflection angle of the object. }
		\label{sigma_add}
	\end{figure}
	
	\clearpage
	%
	%
	\bibliographystyle{splncs04}
	\bibliography{egbib}
	
\end{document}